\documentclass[letterpaper, 10 pt, conference]{ieeeconf}  %

\IEEEoverridecommandlockouts                              %

\usepackage{graphicx}
\usepackage{xcolor}
\usepackage{amsmath}

\usepackage[noadjust]{cite}

\usepackage{hyperref}
\usepackage{amssymb}
\usepackage[font={footnotesize}]{caption}

\usepackage[export]{adjustbox}

\usepackage{booktabs}  %
\usepackage{tabularx}
\newcolumntype{C}{>{\centering\arraybackslash}X}
\usepackage{multicol}  %
\usepackage{multirow}

\usepackage{algpseudocode}
\usepackage{algorithm}
\usepackage{url}

\newcommand{\eg}{\emph{e.g.}}
\newcommand{\ie}{\emph{i.e.}}

\title{\LARGE \bf
Diff-DOPE: Differentiable Deep Object Pose Estimation
}

\author{Jonathan Tremblay, Bowen Wen, Valts Blukis, Balakumar Sundaralingam, Stephen Tyree, Stan Birchfield
\\ NVIDIA%
}

\begin{document}

\twocolumn[{
\renewcommand\twocolumn[1][]{#1}%
\maketitle

\begin{center}
    \centering
    \begin{tabular}{ccc}

        \hspace{-2ex} %
        \includegraphics[trim={200px 140px 100px 270px},clip,width=0.32\textwidth]{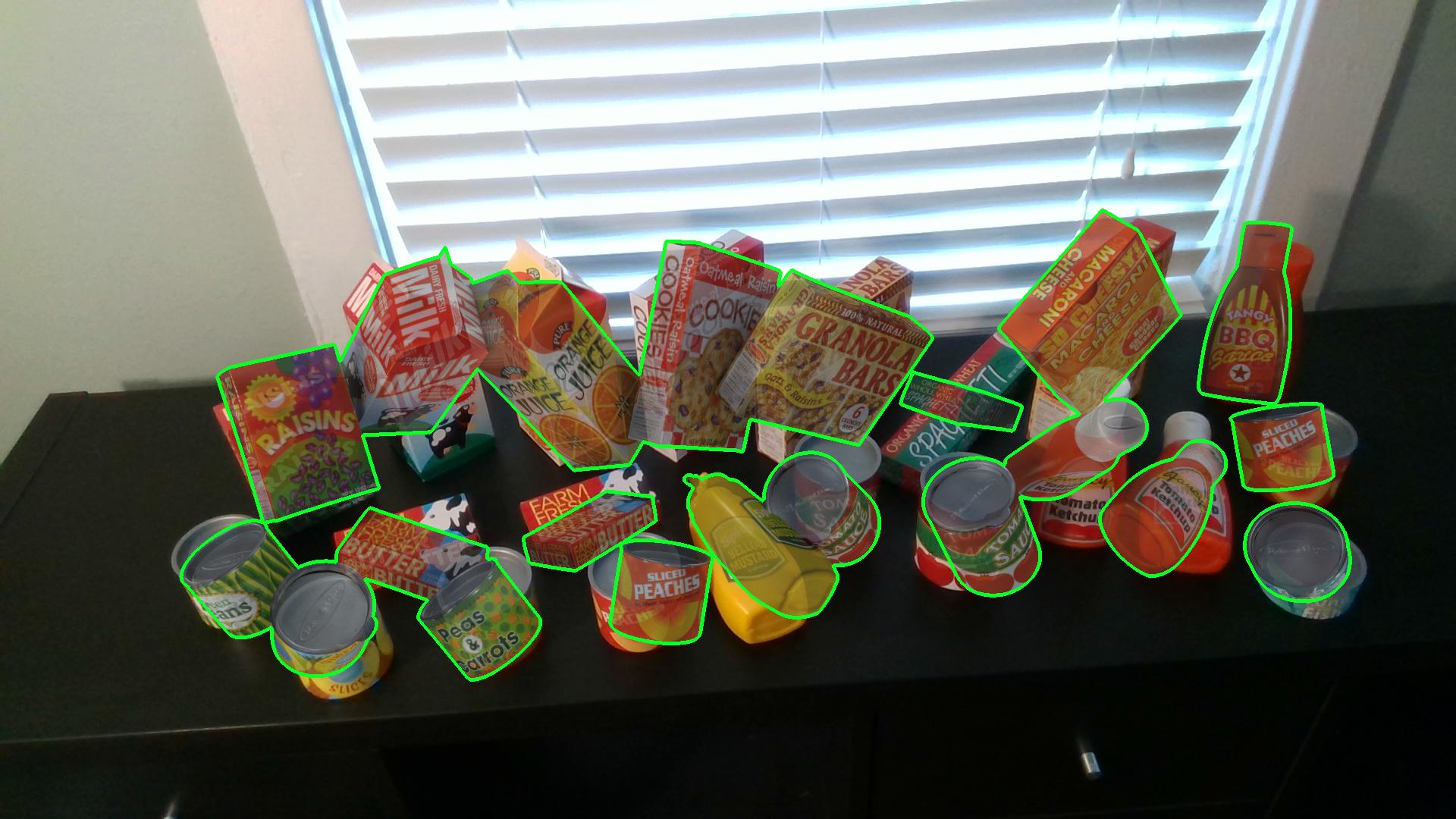} & 
        \hspace{-2ex} %
        \includegraphics[trim={200px 140px 100px 270px},clip,width=0.32\textwidth]{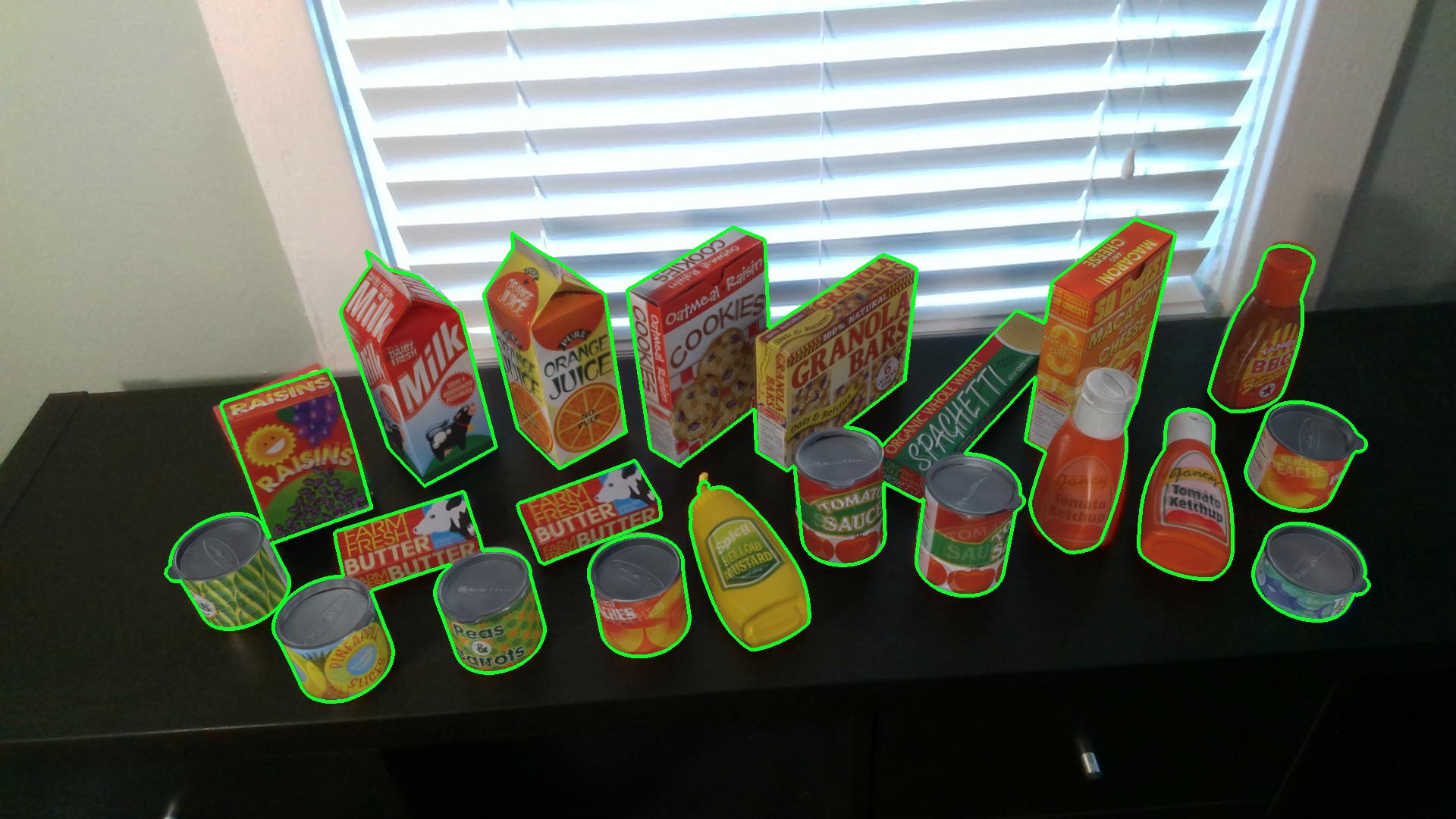} & 
        \hspace{-2ex} %
        \includegraphics[trim={200px 140px 100px 270px},clip,width=0.32\textwidth]{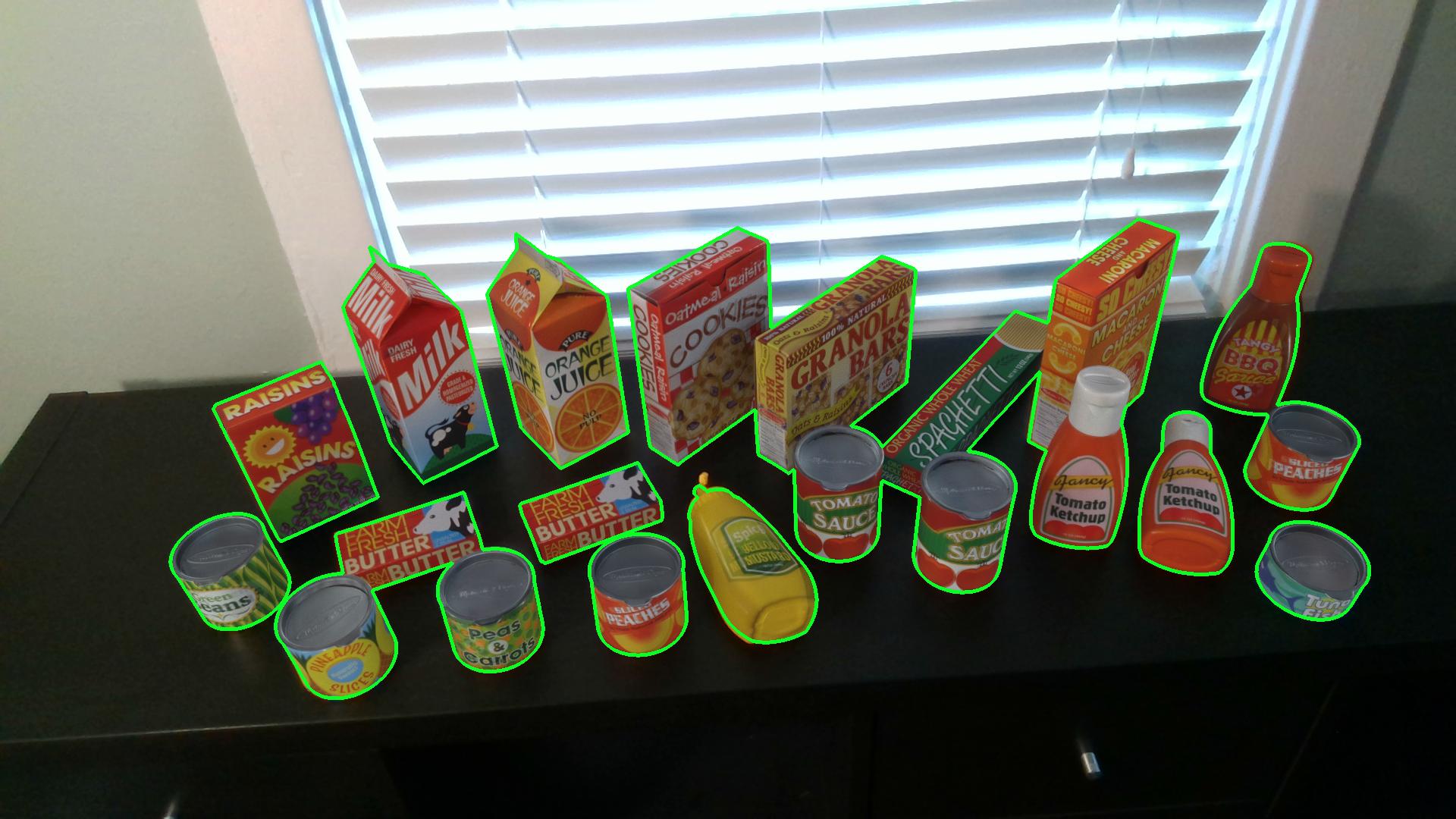} 
        \\
        \hspace{-2ex} %
        \includegraphics[trim={200px 140px 100px 270px},clip,width=0.32\textwidth]{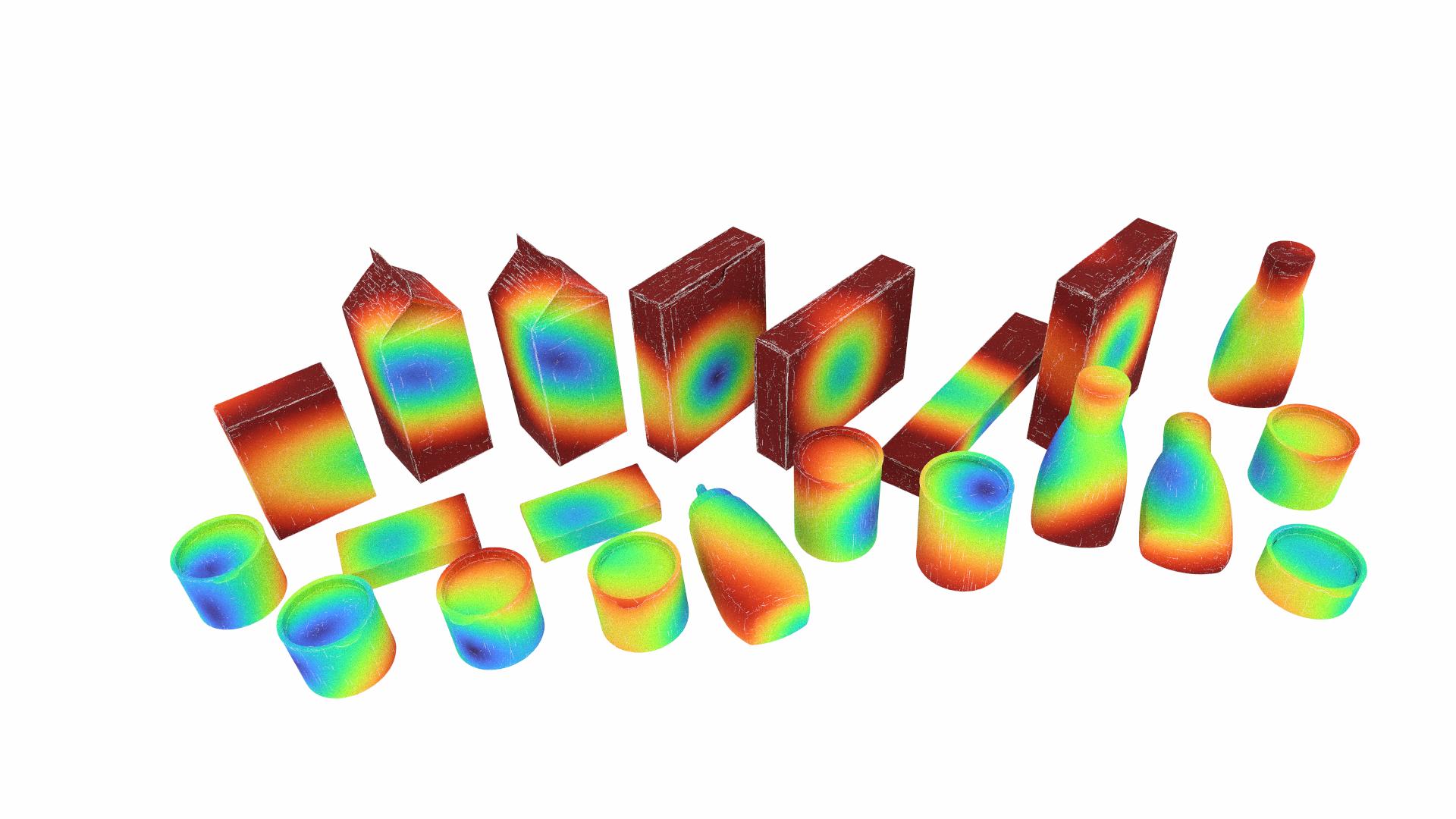} & 
        \hspace{-2ex} %
        \includegraphics[trim={200px 140px 100px 270px},clip,width=0.32\textwidth]{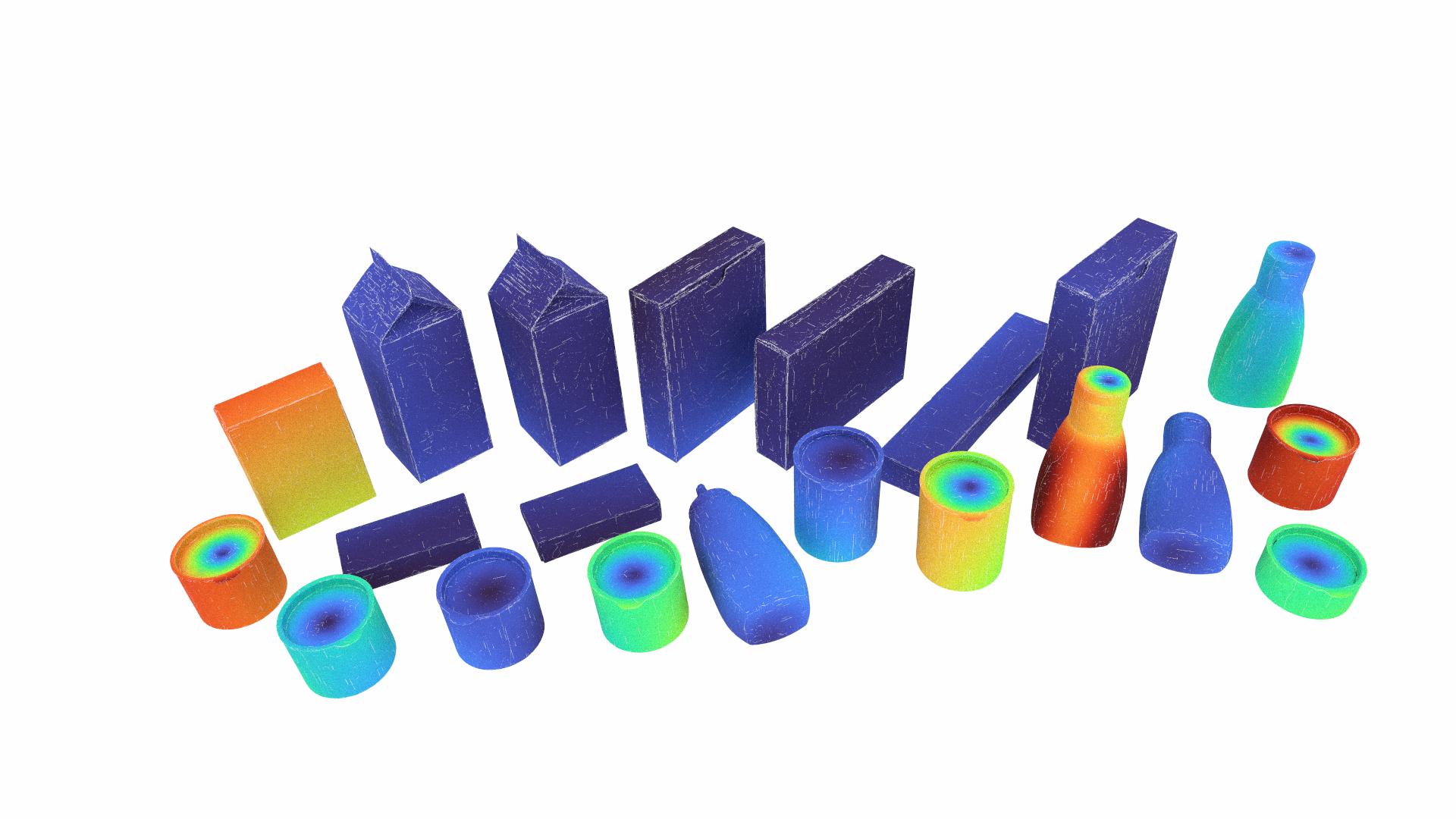} & 
        \hspace{-2ex} %
        \includegraphics[trim={200px 140px 100px 270px},clip,width=0.32\textwidth]{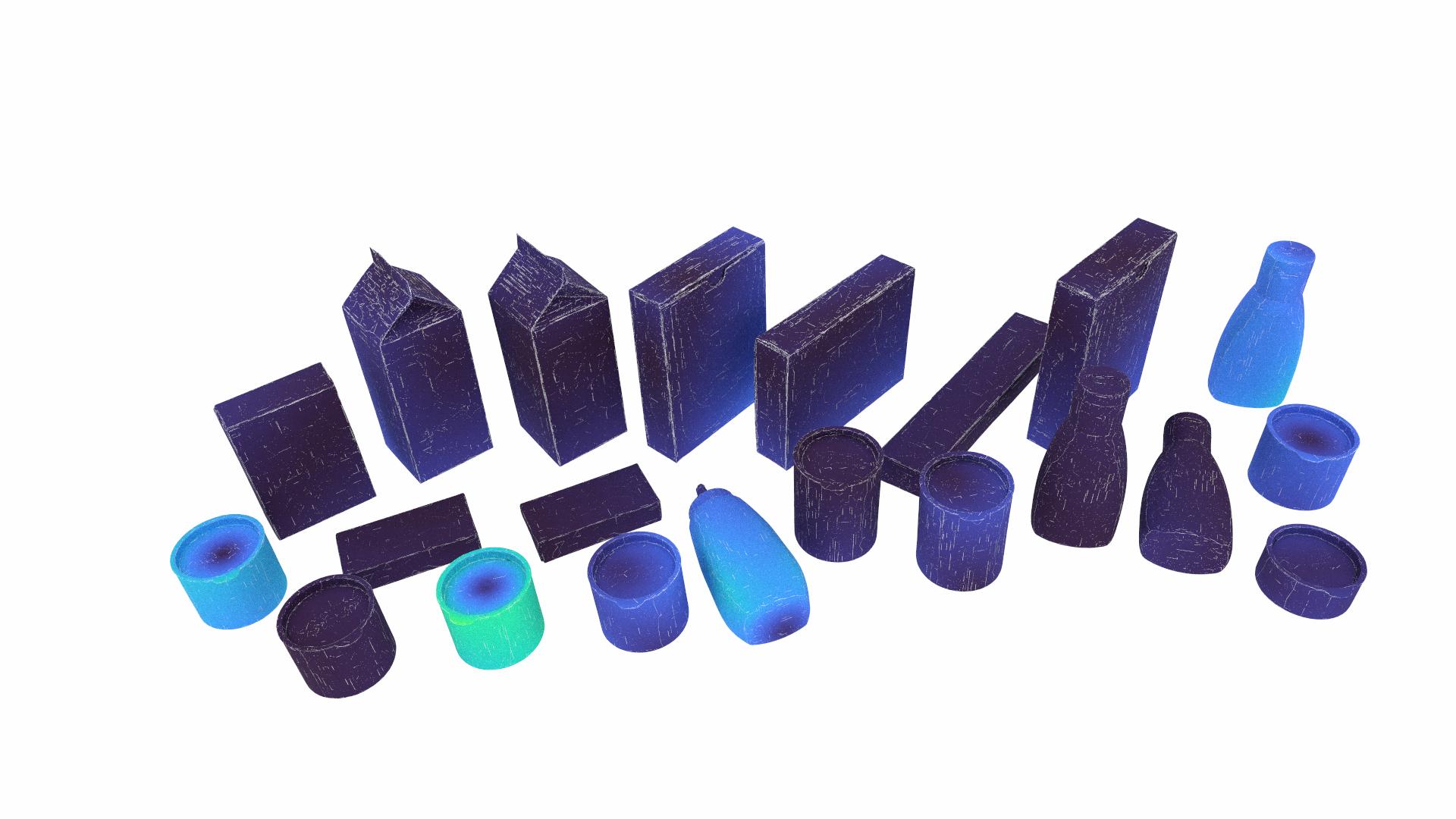} 
        \\
        Input Poses & 
        MegaPose~\cite{labbe2022megapose} & 
        Diff-DOPE (ours) 

    \end{tabular}
    \captionof{figure}{Diff-DOPE uses a differentiable renderer to iteratively refine the 6-DoF pose of an object.  
    Unlike previous approaches, the method works \emph{without any training}.
    Shown are qualitative results on an input scene from the HOPE dataset~\cite{tyree2022iros:hope}. 
    The bottom row shows the error heatmap of each ground truth object, 
    where darker red indicates higher error with respect to the ground truth pose 
    (legend: 0~cm \includegraphics[height=2mm]{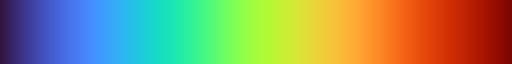} 5~cm).}
    \label{fig:input_im}
\end{center}%
}]

\thispagestyle{empty}
\pagestyle{empty}

\newlength{\mycharwid}  %
\settowidth{\mycharwid}{8}


\begin{abstract}

We introduce Diff-DOPE, a 6-DoF pose refiner that takes as input an image, a 3D textured model of an object, and an initial pose of the object.
The method uses differentiable rendering to update the object pose to minimize the visual error between the image and the projection of the model. 
We show that this simple, yet effective, idea is able to achieve state-of-the-art results on pose estimation datasets.
Our approach is a departure from recent methods in which the pose refiner is a deep neural network trained on a large synthetic dataset to map inputs to refinement steps.
Rather, our use of differentiable rendering allows us to avoid training altogether.
Our approach performs multiple gradient descent optimizations in parallel with different random learning rates to avoid local minima from symmetric objects, similar appearances, or wrong step size. 
Various modalities can be used, \eg, RGB, depth, intensity edges, and object segmentation masks.
We present experiments examining the effect of various choices, showing that the best results are found when the RGB image is accompanied by an object mask and depth image to guide the optimization process.  
The project website is 
\href{https://diffdope.github.io/}{diffdope.github.io}.

\end{abstract}

\section{INTRODUCTION}

Estimating the 6-DoF pose (\ie, six degrees of freedom, including 3D translation and 3D rotation) of an object is 
a crucial task for a wide range of applications, including robotic manipulation, augmented/mixed reality, and autonomous navigation.
Recent years have witnessed remarkable progress in solving this problem at both the instance- and category-level~\cite{tremblay2018corl:dope,lin2022icra:centerpose}, with a variety of scenarios such as per-instance training of networks~\cite{tremblay2018corl:dope,labbe2020eccv:cosypose}, textured 3D object models for comparison at inference time~\cite{labbe2022megapose}, and reference object images~\cite{liu2022eccv:gen6d,he2022neurips:oneposeplusplus}.

Many approaches divide the problem into two stages: first the pose is roughly estimated, then the pose is refined.
This paper focuses on the latter pose refinement step, which is often crucial for obtaining good results.
Classic techniques use the Jacobian between image space and pose space to compute the delta pose needed to minimize image-based error~\cite{pauwels2015simtrack,wen2023bundlesdf,wen2021bundletrack,tjaden2016eccv,tjaden2017iccv}.
More recent work approaches the refinement problem via render-and-compare, where a neural network compares a rendered image of the 3D mesh to the target image to predict an updated pose under which the rendered image better matches the target
\cite{li2018deepim,labbe2020eccv:cosypose,labbe2022megapose,wen2020iros:se3tracknet}.
Because these techniques rely upon a traditional non-differentiable rendering pipeline, the neural network must be trained offline on a large dataset to learn to compute the pose update.
Once trained, the network is opaque, yielding little insight as to why it is performing well or failing, making it difficult to improve performance without costly and time-consuming retraining.

In this work, we leverage recent advancements in differentiable rendering~\cite{laine2020tog:nvdiffrast,liu2019iccv:softras} that make it 
possible to explore the problem of 6-DoF object pose refinement as direct end-to-end optimization.
This approach alleviates the challenge of curating a dataset to train the refiner, and it leads to a solution that is more flexible and interpretable. 
Intuitively, the approach is inspired by the fact that reprojection misalignments produced by even slightly erroneous pose estimates are easily noticeable, which suggests that local pixel information may be enough for high quality pose estimation, even in the absence of learned priors.
Our approach, called \mbox{\emph{Diff-DOPE}} (for \emph{diff}\!erentiable \emph{d}eep \emph{o}bject \emph{p}ose \emph{e}stimation),
is demonstrated in Figure~\ref{fig:input_im}.

The design choice of differentiable rendering allows unprecedented flexibility:  the user can fine-tune the setup to favor certain loss terms tailored to a specific situation without having to retrain a network. 
Our method can be used in a variety of scenarios, \eg, 
RGB only, depth only, or RGBD, as well as being able to optionally consume an object segmentation mask and/or to incorporate intensity edges or other information in the optimization. 
Nevertheless, all our experiments are run with the same parameter values and modalities to demonstrate its broad applicability.

Specifically, our method minimizes the misalignment error of the rendered object with the observed image through gradient descent, in a form of render-and-compare. 
One challenge with such an approach is to avoid local minima during optimization.
To address this problem, we initialize multiple optimization instances in parallel.  
However, rather than randomly perturbing the initial pose as is commonly done, we randomly perturb the learning rates applied to the instances.
This innovation, which we refer to as \emph{learning rate randomization}, is key to the success of our method, as we show in experiments.
On a high-quality dataset like HOPE~\cite{tyree2022iros:hope}, our method achieves better than 1~cm accuracy on complex occluded scenes, oftentimes less than 0.5~cm---which is  better than previous state-of-the-art by a significant margin.
Our method also outperforms previous techniques on the T-LESS~\cite{hodan2017tless} and YCB-Video~\cite{xiang2018rss:posecnn} datasets containing textureless and symmetric objects.

\section{Related Work}
\noindent\textbf{Object pose estimation.}
6-DoF object pose estimation is a well known problem in the robotics and computer vision communities. 
This problem was first approached explicitly, where corresponding points on the 3D object and the image 
were jointly optimized to compute a pose~\cite{pauwels2015simtrack,Lowe1999-bf,bay2006surf,Collet2010-zj,Collet2011-lj,drost2010model}, or where template matching %
correlated object pose to image observations~\cite{Hinterstoisser2011-es,jurie2001simple}.
More recent efforts have focused on learned optimization using a neural network~\cite{Rad2017-de,tremblay2018corl:dope,Kehl2017-ssd,Tekin2017-hp,peng2019pvnet,pavlakos20176,hu2019segmentation,xiang2018rss:posecnn,Park2019-od,song2020hybridpose,zakharov2019dpod,haugaard2021surfemb}.
Some of the best performing methods rely on trainable refinement networks~\cite{wang2019densefusion,labbe2020eccv:cosypose,li2018deepim,lipson2022cvpr:coupitref} which utilize render-and compare~\cite{pauwels2015simtrack,oberweger2015training,manhardt2018deep,labbe2022megapose}. 
These methods iteratively compare rendered images of the object with the observed image via a neural network that outputs a delta pose. 
As the neural network is expected to optimize the similarity in visual appearance of observed and rendered object, 
obscure errors that are hard to explain can appear when objects are significantly different from those in the training set. 
In contrast, we propose to directly leverage the information directly used by render-and-compare, which leads to more predictable performance.

\noindent\textbf{Differentiable rendering.}
Recently, practical methods have been found for differentiable rendering for 
both rasterization~\cite{laine2020tog:nvdiffrast,ravi2020pytorch3d,tensorflow2015-whitepaper,KaolinLibrary} and 
ray tracing~\cite{nimier2019mitsuba,hasselgren2022shape}.
Perhaps the most widely used differentiable renderer is NeRF~\cite{mildenhall2020eccv:nerf}, 
which can be used for camera pose estimation from a novel view~\cite{yen2021inerf,lin2023icra:parallel,wen2023bundlesdf}, 
camera calibration~\cite{jang2021codenerf,levy2023melon},
camera verification~\cite{maggio2023verf},
and object pose estimation~\cite{pan2023learning}.
These methods require high-quality multi-view data with accurate camera poses---requirements that are not always practical in the pose estimation setting.
Perhaps the closest work to ours, EasyHeC~\cite{chen2023easyhec},  uses a similar approach for camera pose estimation by differentiable render-and-compare. 
Their work focuses on robot-to-camera pose estimation and only considers the modality of segmentation masks, whereas we address any object with a 3D textured model and allow for multiple image modalities.

\begin{figure*}[!t]
    \centering
     \includegraphics[width = 1.0\textwidth]{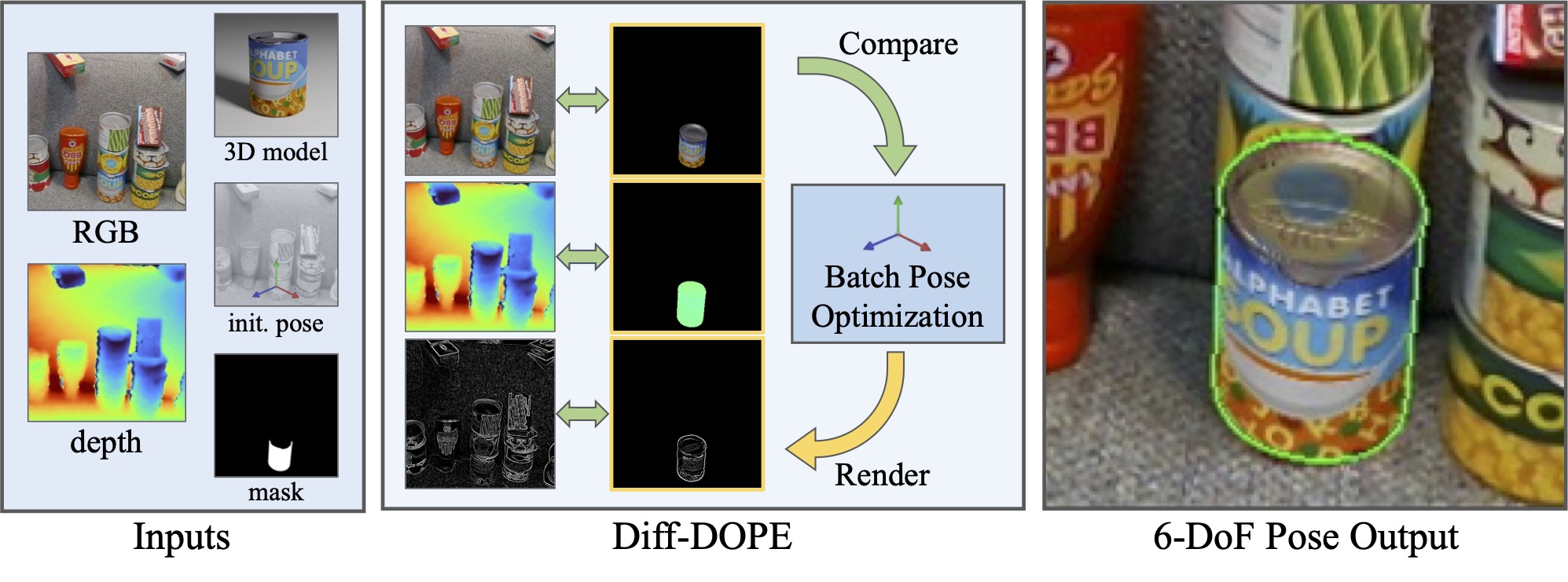}
    \caption{Overview of the Diff-DOPE system, which iteratively refines the 6-DoF pose of an object by minimizing the reprojection error between the rendered 3D model and the input channels (\eg, RGB, depth, and edges).  
    For additional robustness, the algorithm performs multiple optimizations in parallel using randomly sampled initial learning rates;
    once the error stabilizes, the pose with the lowest reprojection error is selected.    
    }\label{fig:overview}
\end{figure*}%

\section{Diff-DOPE}

Formally, the problem of 6-DoF object pose estimation is defined as follows:
Given a 3D textured model $M$ of a rigid object and a camera image $\textit{I}$
where the object appears, find the transformation $T_\textit{CO}$ from the camera to the object, where rotation is represented using a quaternion.
A common paradigm to solve this problem consists of an object detection step followed by a pose refinement step~\cite{wang2019densefusion,labbe2020eccv:cosypose,lipson2022cvpr:coupitref,labbe2022megapose}.
That is, after the object of interest is localized in the image and its initial pose is estimated,
its pose is iteratively updated using a render-and-compare approach.

We propose Diff-DOPE to replace the pose refiner within such a system.
We assume access to a differentiable renderer $\cal R$ which can render a reprojected image ${\hat I} \gets {\cal R}(T_\textit{CO},M,K)$ of the 3D model~$M$ in an arbitrary camera pose $T_\textit{CO}$ with camera intrisics $K$.
We seek to solve the 6-DoF pose estimation problem by minimizing the reprojection error:
\begin{equation}\label{eq:poseproblem}
    T_\textit{CO} = \arg\min_{T_\textit{CO}'} \; \textit{loss}({\cal R}(T_\textit{CO}', M, K), \textit{I})
\end{equation}
where $\textit{loss}$ is an error metric of our design. By adopting a differentiable renderer ${\cal R}$, we have access to gradients computed through the rendering process. This means we can compute gradients w.r.t. the camera pose $T_\textit{CO}$ of a loss function that operates in image-space, allowing a solution to Eq.~\eqref{eq:poseproblem} to be computed by gradient descent.

While it is possible in theory to retrieve object pose by comparing against only an RGB reference image, 
in practice it is non-trivial to render a 3D model so as to account for all lighting and material artifacts in the real scene.
Thankfully, the flexibility of the differentiable rendering framework allows multiple modalities in the loss function.
As such, we compute loss terms that leverage additional information, such as depth, %
edge detection maps, %
and segmentation masks. %
See Fig.~\ref{fig:overview}.

In particular, we define the loss function as a weighted combination of modality-specific terms:
\begin{equation}
\begin{array}{rccl}
    \textit{loss}(\cdot,\cdot) &= 
     & \lambda_c \left| S \odot (I_c - {\cal R}_c(T_\textit{CO},M,K) \right|_1 &+ \\
    && \lambda_d \left| S \odot (I_d - {\cal R}_d(T_\textit{CO},M,K) \right|_1 &+ \\
    && \lambda_e \left| S \odot (I_e - {\cal R}_e(T_\textit{CO},M,K) \right|_1 &
\end{array}
\label{eq:loss}
\end{equation}
We compute the L1 loss between corresponding input images $I_c$, $I_d$, $I_e$ and rendered images ${\hat I}_*\!=\!{\cal R}_*(\cdot,\cdot,\cdot)$, where $* \in \{c,d,e\}$ for RGB/color ($c$), depth ($d$), and edge ($e$) modalities, respectively.
Pixelwise multiplying (\ie, Hadamard product, indicated by $\odot$) by the object mask $S$ limits the loss to relevant regions of the image,
and weights $\lambda_*$ allow loss terms to be balanced or omitted.
Edge maps are computed by applying an off-the-shelf edge detector to the RGB image.

To find the best pose w.r.t. the reprojection error, we use gradient descent without momentum to minimize the loss function in Eq.~\eqref{eq:loss}, as detailed
in Algorithm~\ref{alg:cap}. 
The method takes as input the images $I_c$, $I_d$, $I_e$ and an initial pose $T_\textit{CO}$, as well as the following parameters: the number of iterations $\textit{iters}$, the batch-size \textit{B},
the low learning rate bound ($\ell$), and the high learning rate bound (\textit{h}).
We leverage the parallelism of batch rendering and optimization to run $B$ optimizations concurrently.
Each independent optimization samples a different learning rate from the uniform distribution $U(\ell,h)$ %
and is initialized with the input object pose.
This \emph{learning rate randomization} is similar to the approach of Blier et al.~\cite{blier2018alrao}.
A high learning rate can cause the optimization to find the solution when the initial error is large, 
while a low learning rate is more appropriate when the initial error is small. 
We apply learning rate decay to ensure convergence and prevent oscillation around the final pose.
An alternative approach, similar to \cite{lin2023icra:parallel}, would be to use a fixed learning rate and apply 
noise to the initial pose, but we show that this approach leads to unsatisfactory results.
After the optimizations complete, the pose with the lowest reprojection error is selected among the batch of results.
 
\begin{algorithm}
\caption{Diff-DOPE optimization algorithm}
\label{alg:cap}
\begin{algorithmic}
\Require $T_\textit{CO}, I_c, I_d, I_e, \textit{iters}$ 
\Require $B,\ell,h$ \Comment{{\small batch size, low, high learning rates}}
\For{$i \gets 1$ to $B$} \Comment{\small process batch instances in parallel}
\State $\alpha_i \gets U(\ell,h)$ %
\Comment{\small learning rate randomization}
\State $T_{\textit{CO},i} \gets T_\textit{CO}$ %
\For{$k \gets 1$ to $\textit{iters}$}
\State $\textit{loss}_i \gets$ compute Eq.\eqref{eq:loss} with $T_{\textit{CO},i}$
\State $\alpha \gets \alpha_i \cdot 0.1^{k/\textit{iters}}$
\State $T_{\textit{CO},i} \gets \textit{GradientDescent}(\textit{loss}_i, \alpha)$
\EndFor
\EndFor
\State $j \gets \arg \min_i \hspace{1mm} \textit{loss}_i \hspace{5mm}  i = 1, \ldots, B$
\State $T_\textit{CO} \gets T_{\textit{CO},j}$
\end{algorithmic}
\end{algorithm}

\section{EXPERIMENTAL RESULTS}

In this section we explore the performance of Diff-DOPE on several standard pose estimation datasets, as well as 
investigate the influence of various hyperparameters on performance. 

\subsection{Implementation details and performance}

Diff-DOPE is implemented in Python with PyTorch bindings. 
All experiments were run on an NVIDIA 4090 GPU. 
We use the open-source implementation for differentiable Canny edge detection from Kornia~\cite{eriba2019kornia}; we leave as future work 
modern edge detection algorithms or other image filters. 
For all experiments in this section we used the same hyperparameters.
All weights were set to 1 for the loss in Eq.~\eqref{eq:loss}, 
the batch size was $B=32$, the optimization was run for 100 iterations, 
 the learning rate bounds were $\ell=0.001$ and $h=50$, and the learning rate decay was $0.1$.

Using the functions offered by Nvdiffrast~\cite{laine2020tog:nvdiffrast}, we implemented our own depth map exporter.
We also implemented a CUDA kernel dedicated to matrix multiplication 
that allows
gradient flow for 3D point transformations (from local space to camera space). %
Compared to the na\"ive PyTorch version, this implementation yields a 40x performance improvement, \eg, from 27 ms to 0.63 ms for forward and backward multiplication on a set of 262k points.
Given the hyperparameters above and the matrix multiplication CUDA code, 
Diff-DOPE takes up to 3.5 seconds to optimize the pose of an object, with each iteration taking 34~ms.

\begin{figure*}[!t]
    \centering
    \setlength{\tabcolsep}{0pt}
    \begin{tabularx}{\textwidth}{CCCp{0.03\textwidth}CCCp{0.03\textwidth}CCC}
        \multicolumn{3}{c}{Easy} && 
        \multicolumn{3}{c}{Medium} && 
        \multicolumn{3}{c}{Hard} \\
        \multicolumn{3}{c}{\includegraphics[width=0.31\textwidth]{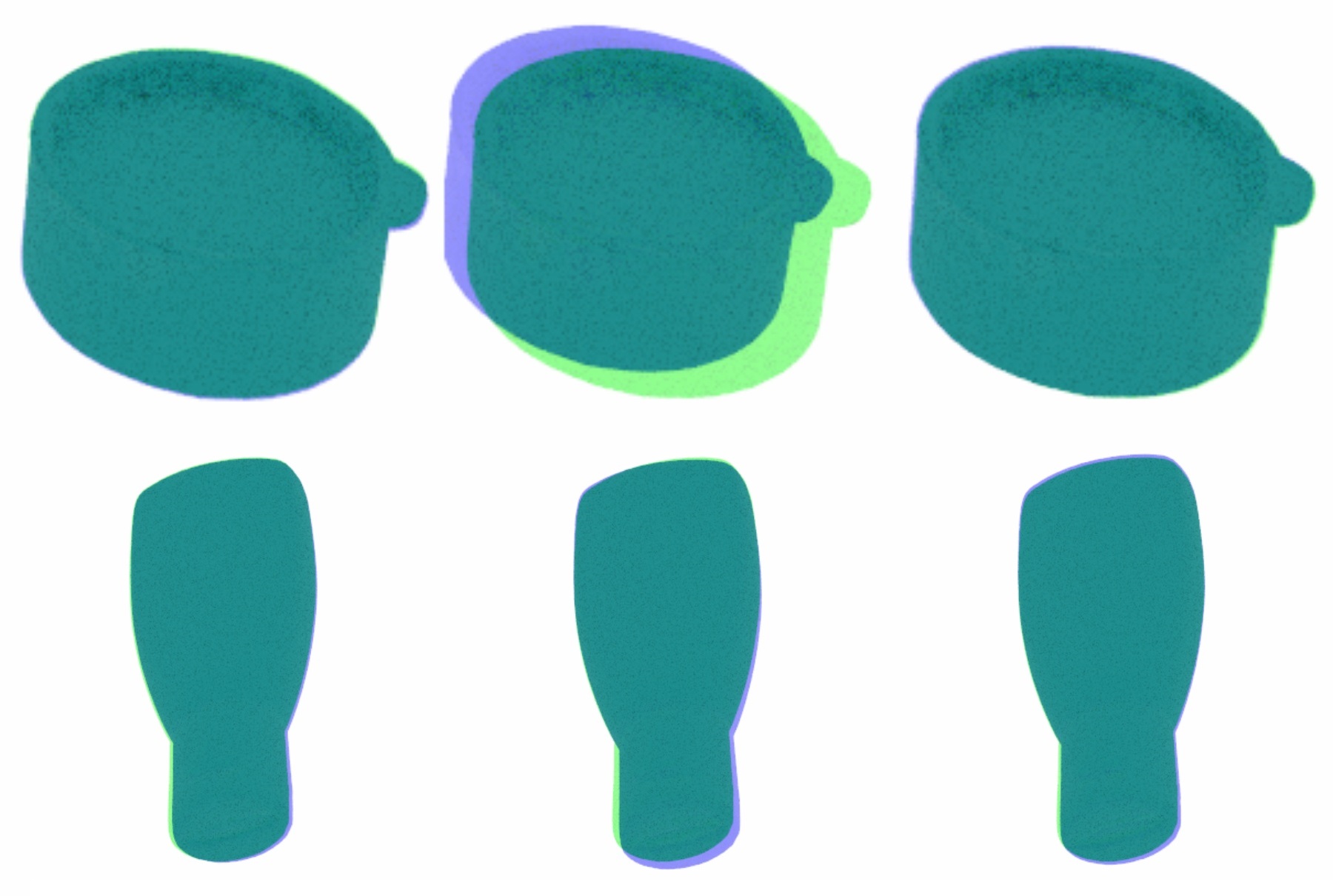}} &&
        \multicolumn{3}{c}{\includegraphics[width=0.31\textwidth]{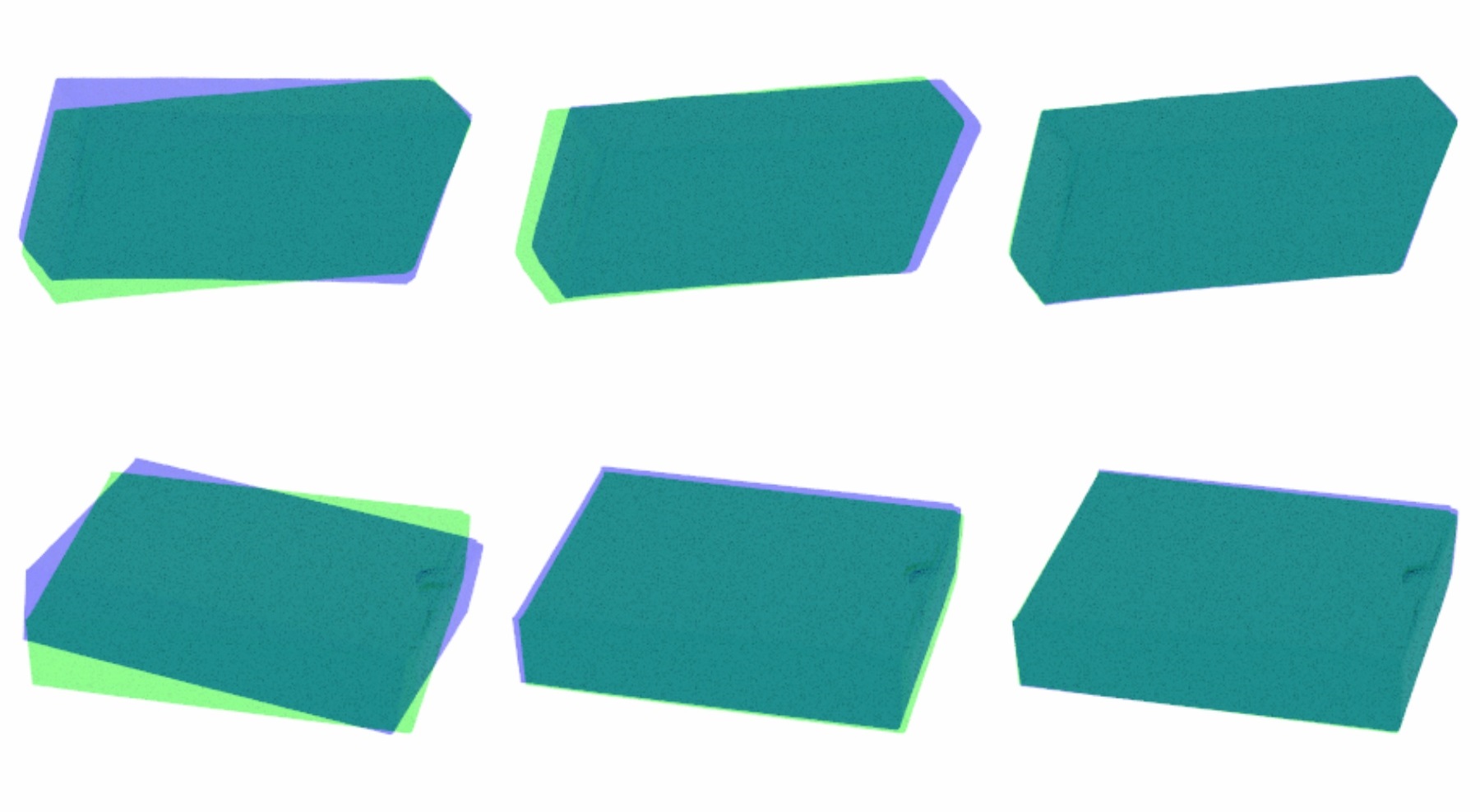}} &&
        \multicolumn{3}{c}{\includegraphics[width=0.31\textwidth]{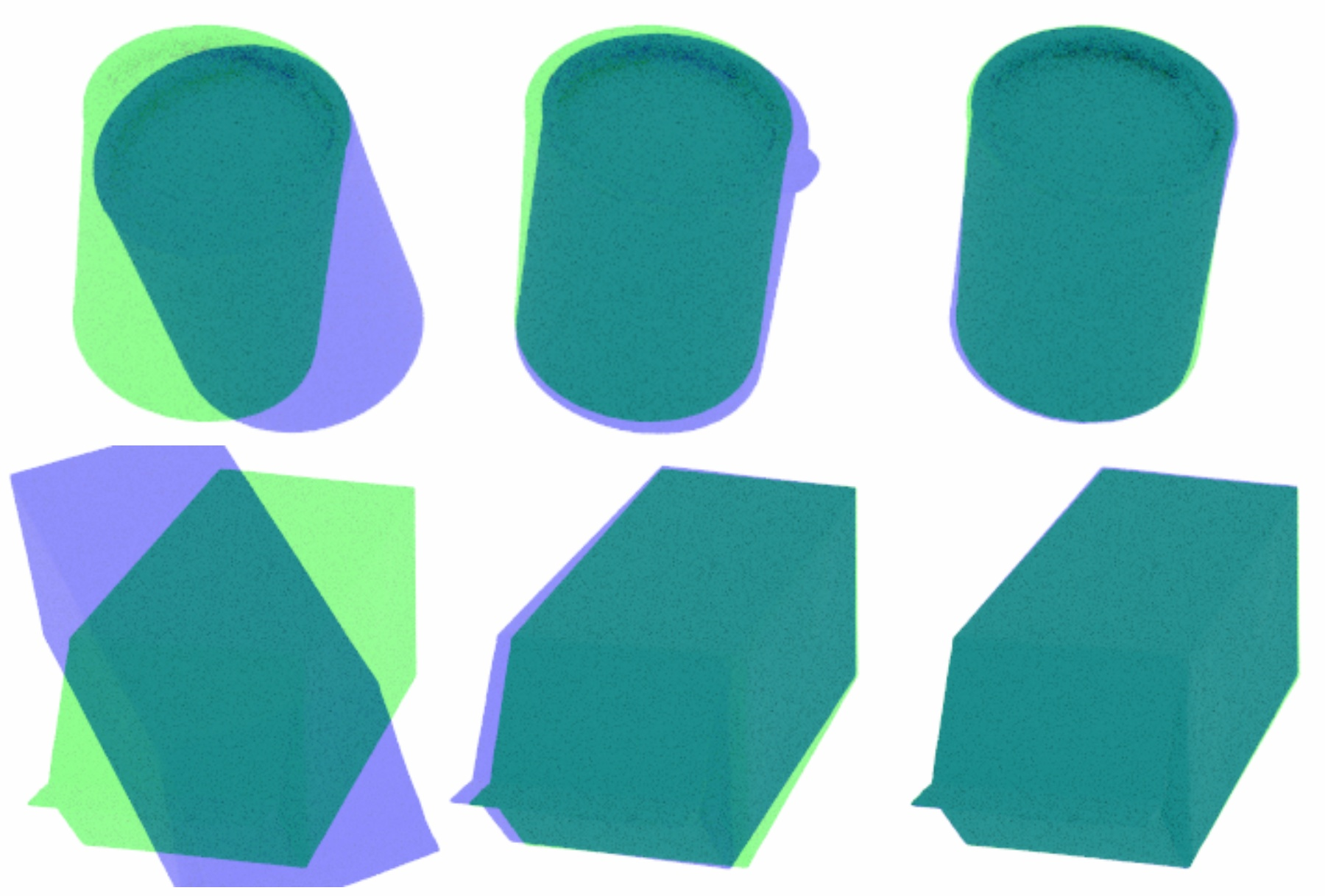}} \\
        {\small Input Pose} & {\small MegaPose} & {\small Diff-DOPE} &&
        {\small Input Pose} & {\small MegaPose} & {\small Diff-DOPE} &&
        {\small Input Pose} & {\small MegaPose} & {\small Diff-DOPE} \\
    \end{tabularx}
    \caption{
    The three different input noise levels of ``easy'', ``medium'', and ``hard'' for sample objects from the  HOPE dataset~\cite{tyree2022iros:hope}, along with qualitative results of Diff-DOPE compared with MegaPose~\cite{labbe2022megapose}.
    Ground truth is shown in green, predictions in purple; teal (green-blue) indicates overlap between the two.   
    }
    \label{fig:megapose_vs_diffdope}
\end{figure*}

\subsection{Accuracy}
\noindent\textbf{Datasets \& metrics}. To evaluate Diff-DOPE as a refiner, 
we create initial poses by applying varying levels of noise to the ground-truth poses, and evaluate our method's ability to recover the correct pose.
To sample the initial poses, for each object in each scene we pick two random axes (one for rotation and the other for translation) and 
apply one of the following rotation and translation perturbations: 
($1^\circ$, 0.1~cm), ($10^\circ$, 1~cm), or ($40^\circ$, 2~cm). 
Figure~\ref{fig:megapose_vs_diffdope} shows some examples of these noise levels, which we refer to as ``easy'', ``medium'', and ``hard'', respectively.
We seek to understand the behavior of our method on small and large input errors, \eg, can our method recover from large pose errors? And, if the initial guess is already good, does our method still improve it or make it worse?

We conducted experiments on the following object pose estimation  datasets:  HOPE~\cite{tyree2022iros:hope}, T-LESS~\cite{hodan2017tless}, and YCB-Video~\cite{xiang2018rss:posecnn}.
For each dataset, we selected the first image from 10 random scenes, and 
applied the three noise levels as a proxy for different initial pose errors.
Our approach is compared with  MegaPose~\cite{labbe2022megapose}, a leading object pose refiner that improves upon CosyPose~\cite{labbe2020eccv:cosypose} and DeepIM~\cite{li2018deepim}.
MegaPose was used with the provided weights for RGBD inputs, while omitting  
the coarse pose estimator in order to focus on the pose refiner.

To evaluate performance, we use the ADD metric~\cite{xiang2018rss:posecnn,tremblay2018corl:dope} which measures the average Euclidean distance between a set of points on the object at the ground-truth pose and at the output pose.
This metric is motivated by its geometric interpretability, a quality particularly important in the context of robotics.
For T-LESS only, we use the symmetric version, ADD-S, since many of the objects exhibit symmetries.
We present threshold plots that show the percentage of predicted poses with error under a specific ADD (or ADD-S) value. 
Such plots are summarized by a single number using the area under the curve (AUC) up to a maximum threshold (5~cm unless specified otherwise).  

\begin{figure}[!t]
    \setlength{\tabcolsep}{0pt}
    \centering
    \begin{tabular}{ccc}
        \includegraphics[trim={5px 5px 5px 5px},clip,width=0.24\textwidth]{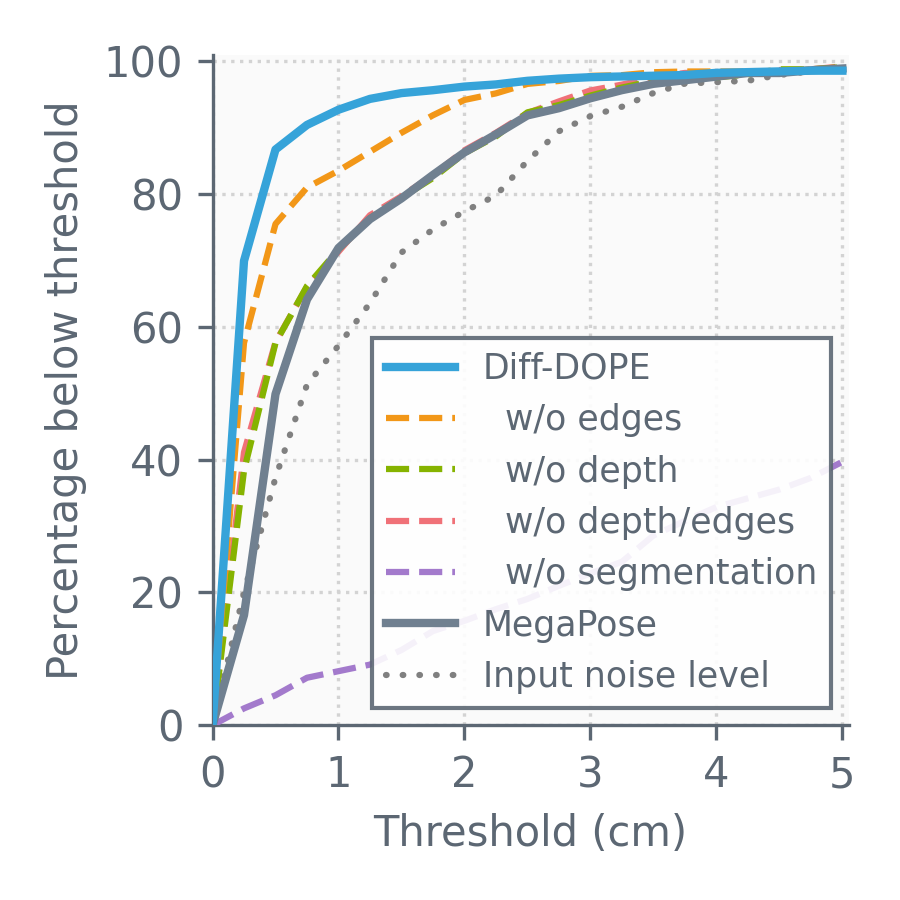} & 
        \includegraphics[trim={5px 5px 5px 5px},clip,width=0.24\textwidth]{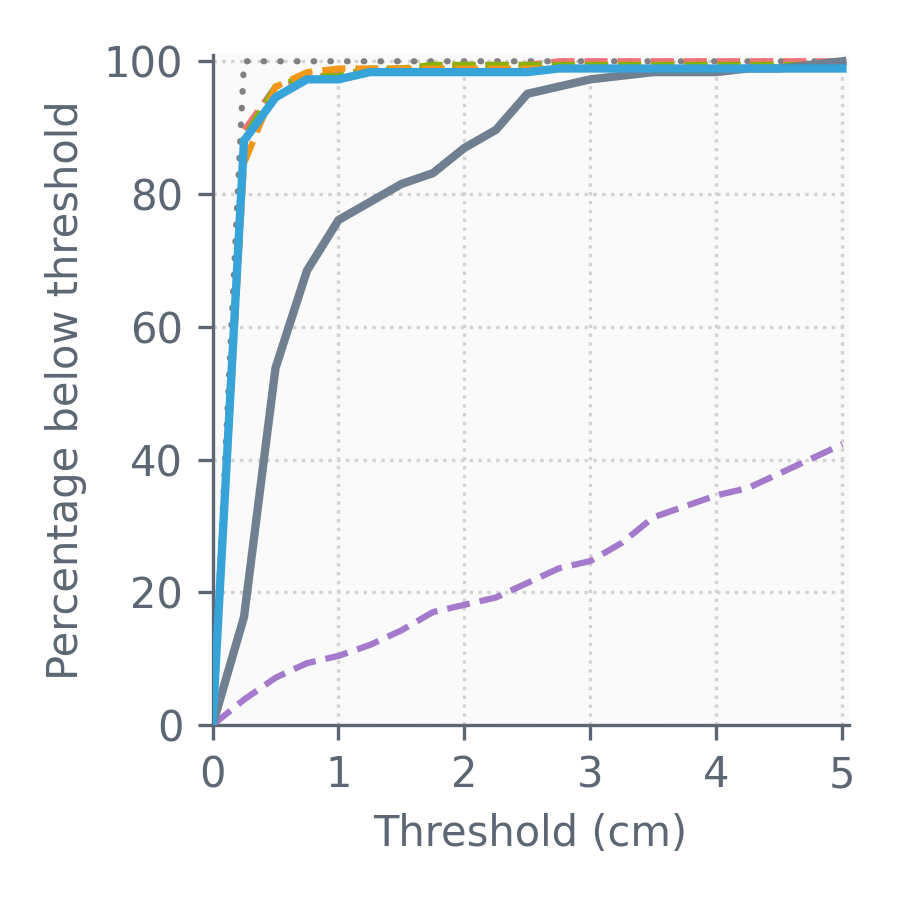}\\
        \includegraphics[trim={5px 5px 5px 5px},clip,width=0.24\textwidth]{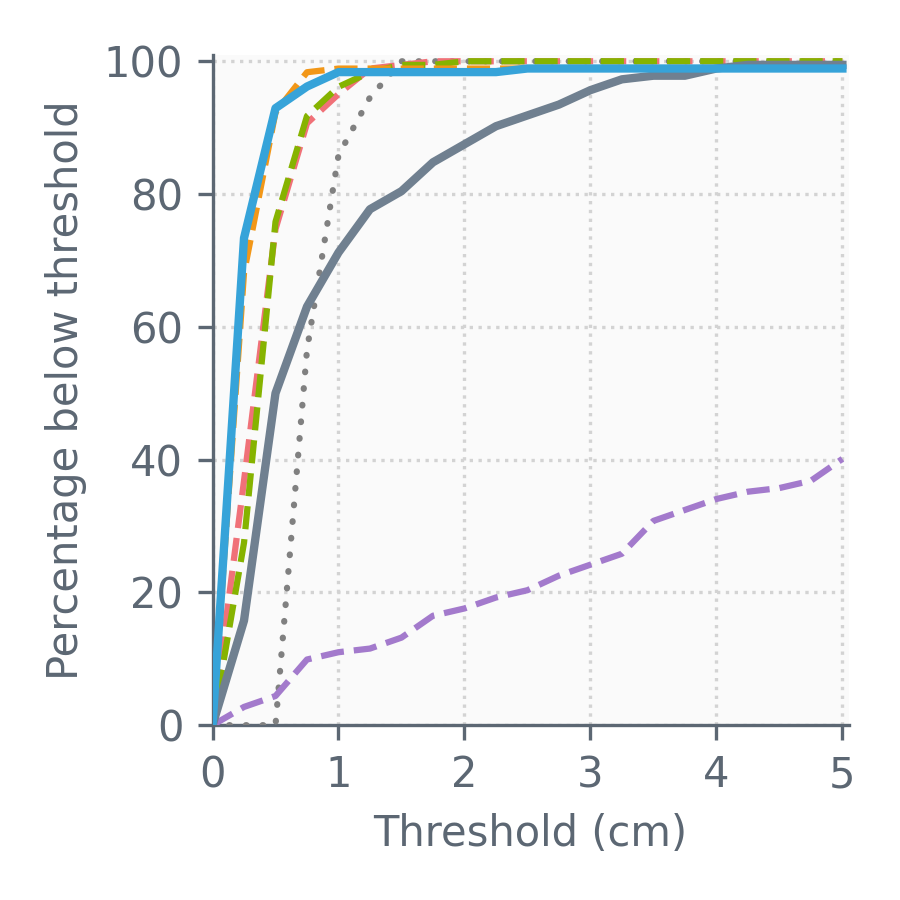} & 
        \includegraphics[trim={5px 5px 5px 5px},clip,width=0.24\textwidth]{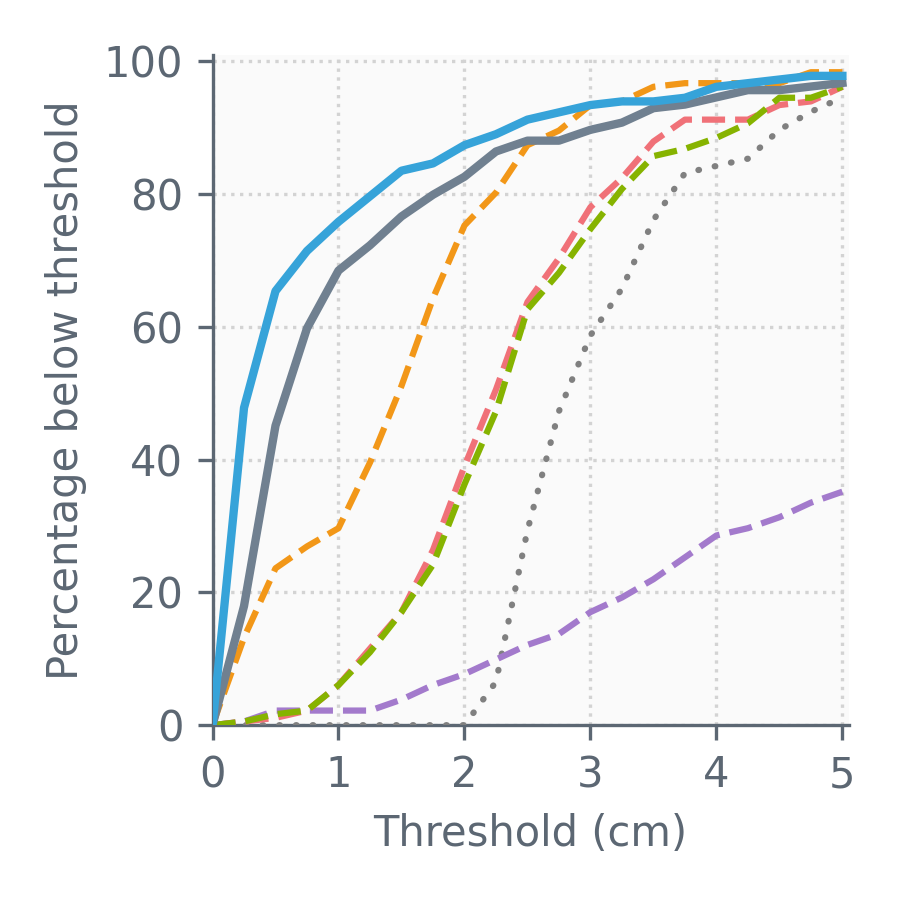}\\
        
    \end{tabular}
    \caption{ADD threshold curves for MegaPose~\cite{labbe2022megapose} and different variations of Diff-DOPE, \eg, without depth; also shown is the ADD error curve for the noisy input poses.
    The subplots show results for all input poses (top-left), and the subset of easy (top-right), 
    medium (bottom-left), and hard (bottom-right) input poses.}
    \label{fig:add_grid}
\end{figure}

\noindent\textbf{Accuracy on HOPE}.
Qualitative results of Diff-DOPE versus MegaPose were shown in Figures~\ref{fig:input_im} and~\ref{fig:megapose_vs_diffdope} for various images from the HOPE dataset.
For quantitative results, 
Figure~\ref{fig:add_grid} compares Diff-DOPE and the MegaPose refiner with different amounts of noise applied 
to ground truth, as described above. 
We observe that Diff-DOPE is more robust to the input noise than MegaPose.
These plots also reveal that segmentation contains the most important information (beyond the RGB input), followed by depth, then edges.
Because the ``easy'' inputs are so close to ground truth, both methods slightly increase the error, although
Diff-DOPE actually improves half of these poses. 
Another interesting observation is that MegaPose performs approximately the same regardless of input noise level, whereas Diff-DOPE's performance drops noticeably on the ``hard'' poses.
For more details, see the AUC numbers in Table~\ref{tab:results}.

\begin{table}
    \centering
    \caption{Comparison of Diff-DOPE (DD) and MegaPose (MP)~\cite{labbe2022megapose} on three datasets using area under the curve (AUC) up to a 5~cm threshold for the ADD (or ADD-S$^\dagger$) error metric.}
    \begin{tabular}{cccccc}
    \toprule
    dataset & method & easy & medium & hard & all \\
    \midrule
         & Diff-DOPE   & \textbf{95.35} & \textbf{94.56} & \textbf{84.03} & \textbf{92.30} \\
    HOPE & MegaPose    & 83.17 & 82.07 & 78.12 & 81.35 \\
         & Input error & 97.50 & 84.35 & 38.29 & 76.17 \\
    \midrule
           & Diff-DOPE & 93.93 & 91.09 & 74.13 & 86.38 \\
    T-LESS$^\dagger$ & DD w/o RGB & \textbf{96.38} & \textbf{92.15} & 74.32 & \textbf{87.42} \\
           & MegaPose & 86.05 & 85.86 & \textbf{85.94} & 85.90 \\
    \midrule
          & Diff-DOPE & \textbf{94.40} & \textbf{89.63} & 65.04 & \textbf{83.03} \\
    YCB-V & MegaPose & 81.81 & 82.31 & 79.27 & 81.13 \\
          & MP $\rightarrow$ DD & 84.73 & 81.86 & \textbf{79.41} & 82.00 \\
    \bottomrule
    \end{tabular}
    \label{tab:results}
\end{table}

\begin{figure}[!t]
    \centering
    \setlength{\tabcolsep}{0pt}
    \begin{tabular}{cc}
        \includegraphics[width=0.24\textwidth]{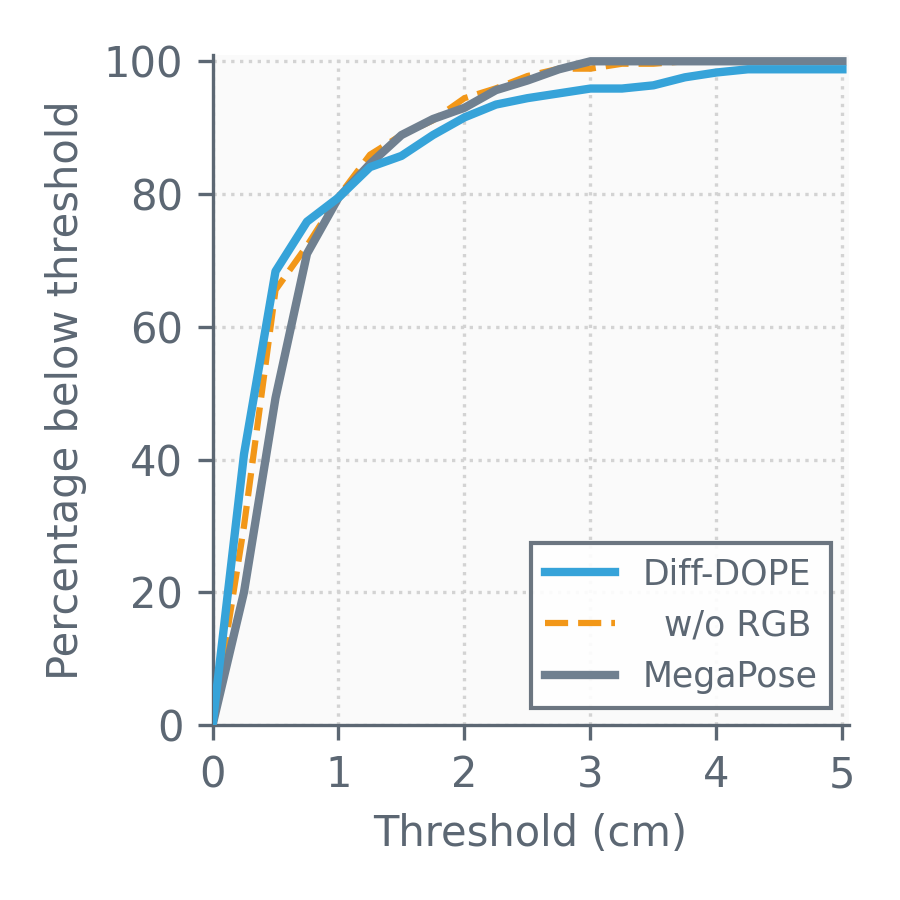} &
        \includegraphics[width=0.24\textwidth]{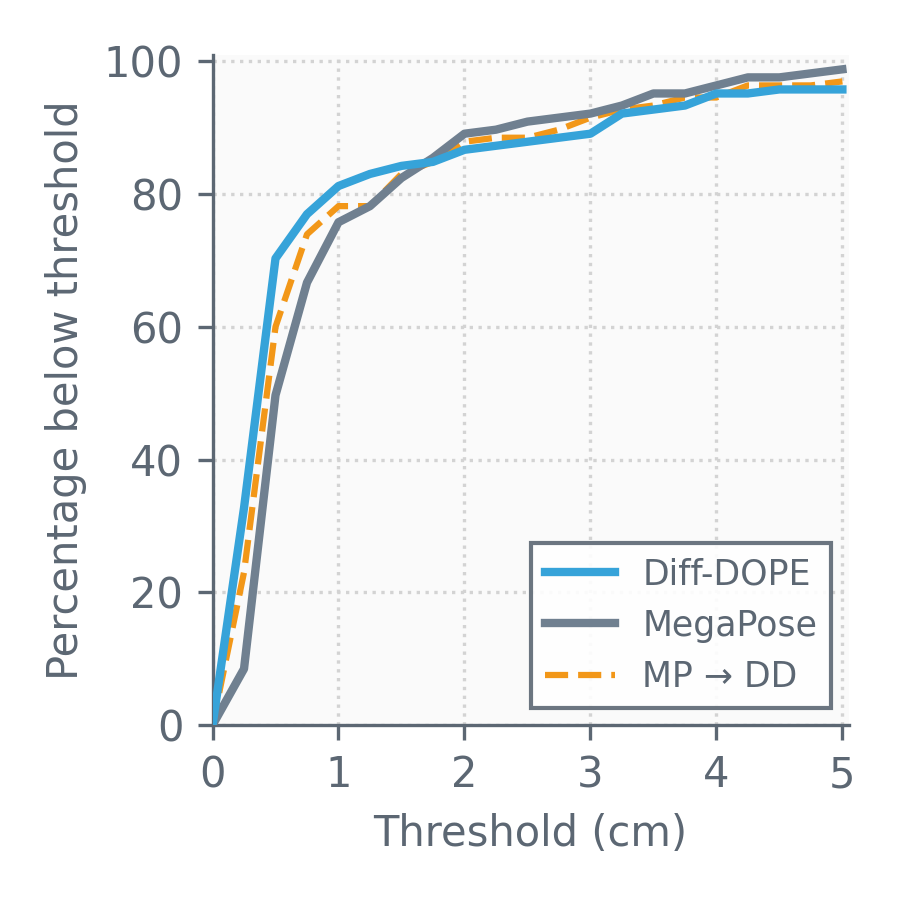}
        
    \end{tabular}
    \caption{
        Left: Results on T-LESS dataset~\cite{hodan2017tless} using ADD-S, 
        where we also evaluate our method without RGB inputs. 
        Right: Results on YCB-Video dataset~\cite{xiang2018rss:posecnn} using ADD, where we also apply our method to the output of MegaPose, which generally helps results at lower input noise levels.
    }
    \label{fig:tless_ycbv_curves}
\end{figure}

\begin{figure}[!t]
    \centering
    \setlength{\tabcolsep}{0pt}
    \begin{tabular}{cccc}
        \includegraphics[width=0.125\textwidth]{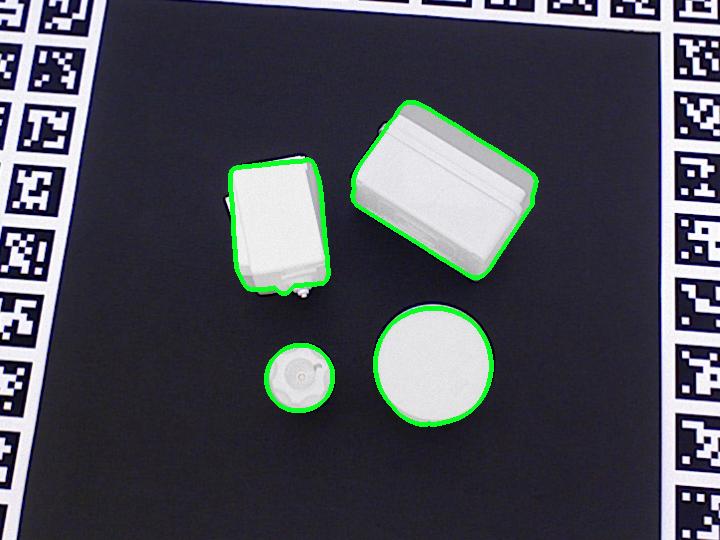} &
        \includegraphics[width=0.125\textwidth]{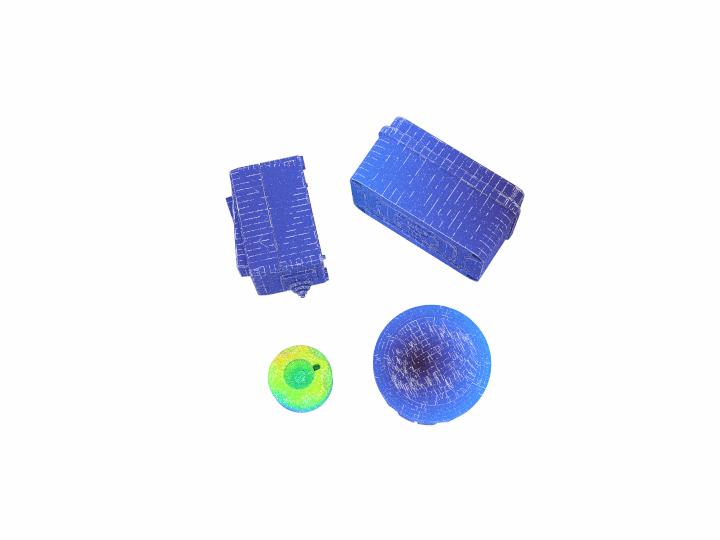} 
        &
        \includegraphics[width=0.125\textwidth]{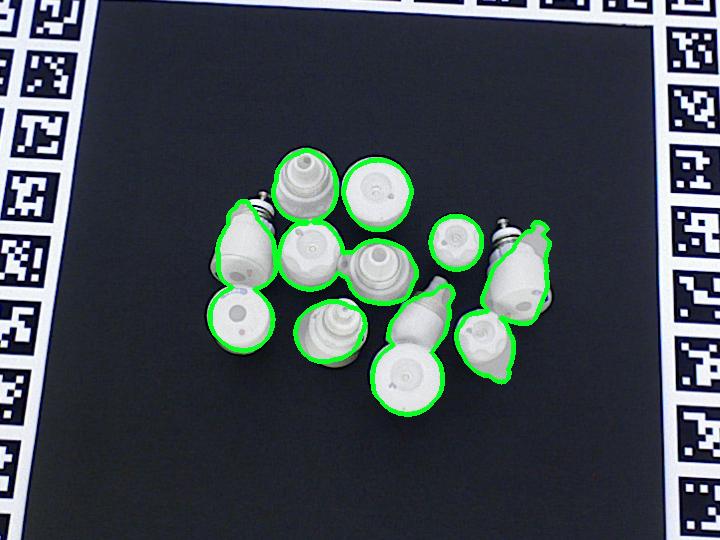} &
        \includegraphics[width=0.125\textwidth]{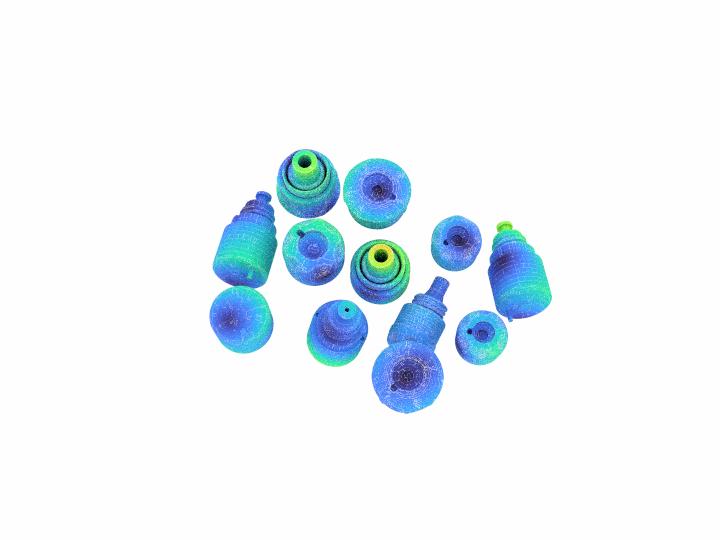} 
        \\
        \includegraphics[width=0.125\textwidth]{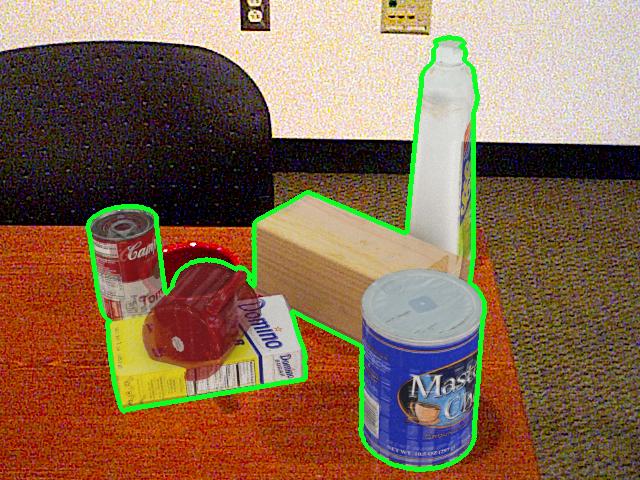} &
        \includegraphics[width=0.125\textwidth]{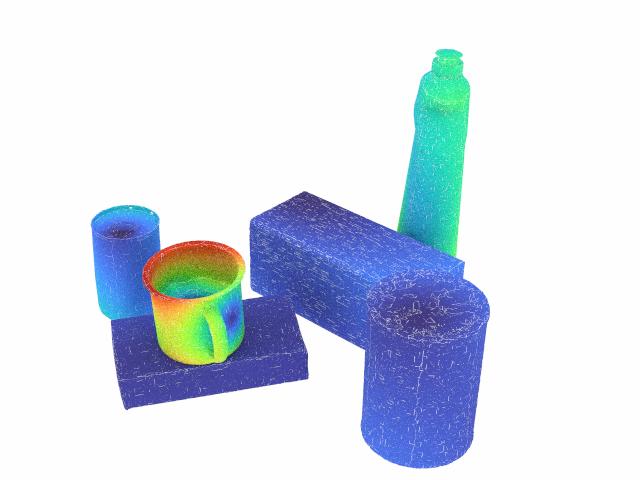} 
        &
        \includegraphics[width=0.125\textwidth]{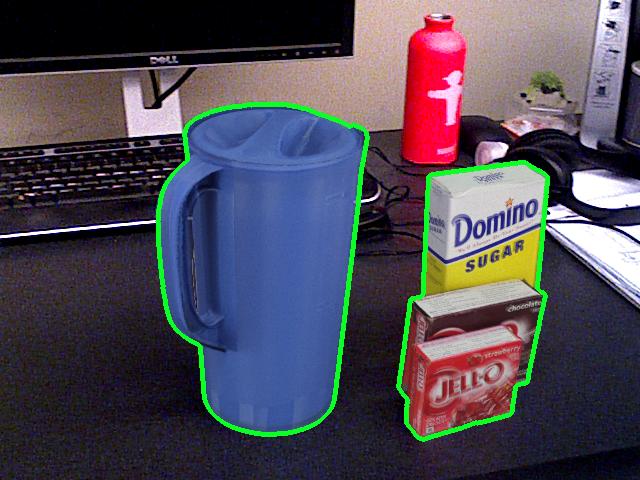} &
        \includegraphics[width=0.125\textwidth]{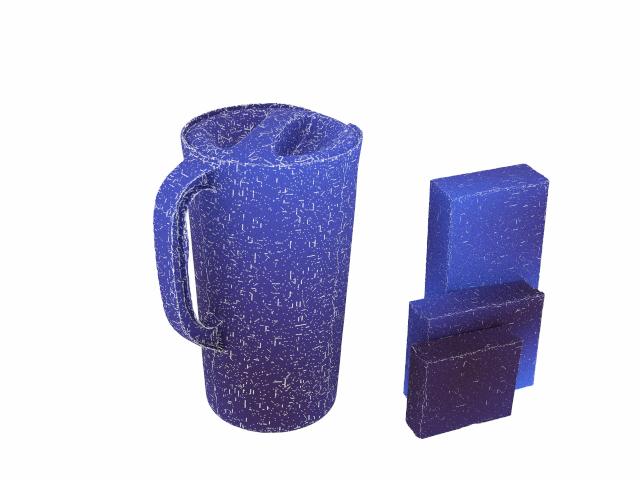} 
        
    \end{tabular}
    \caption{
    Qualitative results of Diff-DOPE on hard scenes from
     T-LESS~\cite{hodan2017tless} (first row) and YCB-Video~\cite{xiang2018rss:posecnn} (second row). Shown are the overlaid poses (left) and the error heatmap
    (right; legend: 0~cm \includegraphics[height=2mm]{figures/turbo.jpg} 5~cm).
    }
    \label{fig:tless_ycbv}
\end{figure}

\noindent\textbf{Accuracy on T-LESS and YCB-Video}.
The HOPE dataset above has high-quality textured objects and images, making it a good fit for our Diff-DOPE method. 
Here we explore the performance of the method when applied to lower-quality datasets, namely, T-LESS (which exhibits poor object reconstruction) and YCB-Video (which contains rather noisy RGB images).
For both datasets we generated similar noisy initial poses as previously shown using ground truth annotation.
Figure~\ref{fig:tless_ycbv_curves} shows quantitative results for both datasets, and Figure~\ref{fig:tless_ycbv} qualitative results. 

The T-LESS dataset provides various 3D assets. 
We use the ``semi-automatically 
reconstructed models which also include surface color''~\cite{hodan2017tless}, 
as the provided CAD models have no surface colors.
Our results are generally better than those of MegaPose for smaller input noise levels, and slightly better overall, but they are worse for  larger input errors.
In the process of analyzing the results we noticed that the provided textured models do not always align well with 
the RGB images, \eg, exaggerated round edges or mismatched colors.
Therefore, we also explored using only the segmentation as the signal instead of matching the RGB colors, which improved results noticeably. 

Even though YCB-Video has well-textured 3D assets, the RGB image quality is less than ideal, with color saturation and heavy noise patterns. 
As a result, all methods perform worse on this dataset.
As before, Diff-DOPE outperforms MegaPose at lower input errors, but performs worse at higher input errors.
In this setting, using only segmentation actually decreases accuracy (from 83.03 to 80.40 AUC), which is expected since YCB-Video is a richly textured dataset.  By comparison, Megapose achieves 81.13 AUC.
We also verified on this dataset that Diff-DOPE can be used to improve the MegaPose prediction by further refining the output of MegaPose.
However, although the results are improved compared with MegaPose (82.00 versus 81.13), the AUC is still lower than running Diff-DOPE directly (83.03). 

These results also allow us to compare the impact of 3D models versus image quality. 
Given the AUC of 92.30 for HOPE, we see that the performance drops the most for YCB-Video (to 83.03), indicating that degradation in the RGB signal is extremely impactful.  
To a lesser extent, lower quality meshes affect results, as seen in the results for T-LESS, where we observe a less dramatic decrease in performance (86.38).

\noindent\textbf{Ablations}.
To better understand the algorithm, we explored variations of the method presented in Algorithm~\ref{alg:cap}.
(The numbers do not match those of the table, due to using a different AUC threshold.)
By default, Diff-DOPE performs uniform sampling of the learning rate between two bounds; we refer to this method as \textit{uniform}.
If one bound is small, \eg,
0.01, and the upper bound is larger, \eg, 10, most samples taken will be skewed towards larger values. 
An alternate approach is to sample the log scale uniformly, which we refer to as \textit{exponent}; this approach skews samples towards 
small values. 
When the input error is small (0.1~cm) \textit{exponent} (95.8 AUC) performs slightly better than \textit{uniform} (95.5 AUC). 
When the input error is larger (2~cm), the opposite occurs: \textit{uniform} (90.7) outperforms \textit{exponent} (89.3).
This result is expected, because with large error the pose update needs to be larger, which is facilitated by higher learning rates.

An alternative to sampling different learning rates, as we do, would be to simply add more pose noise to the input pose to allow the parallel optimizations to explore more of the space.  
When we compare this idea to that of sampling different learning rates,  
we see a significant decrease in performance. 
On ``easy'' data, for example, the AUC drops catastrophically from 83.3 to 57.8.

\begin{figure}[!t]
    \centering
    \setlength{\tabcolsep}{0pt}
    \begin{tabular}{cc}
        \includegraphics[width=0.24\textwidth]{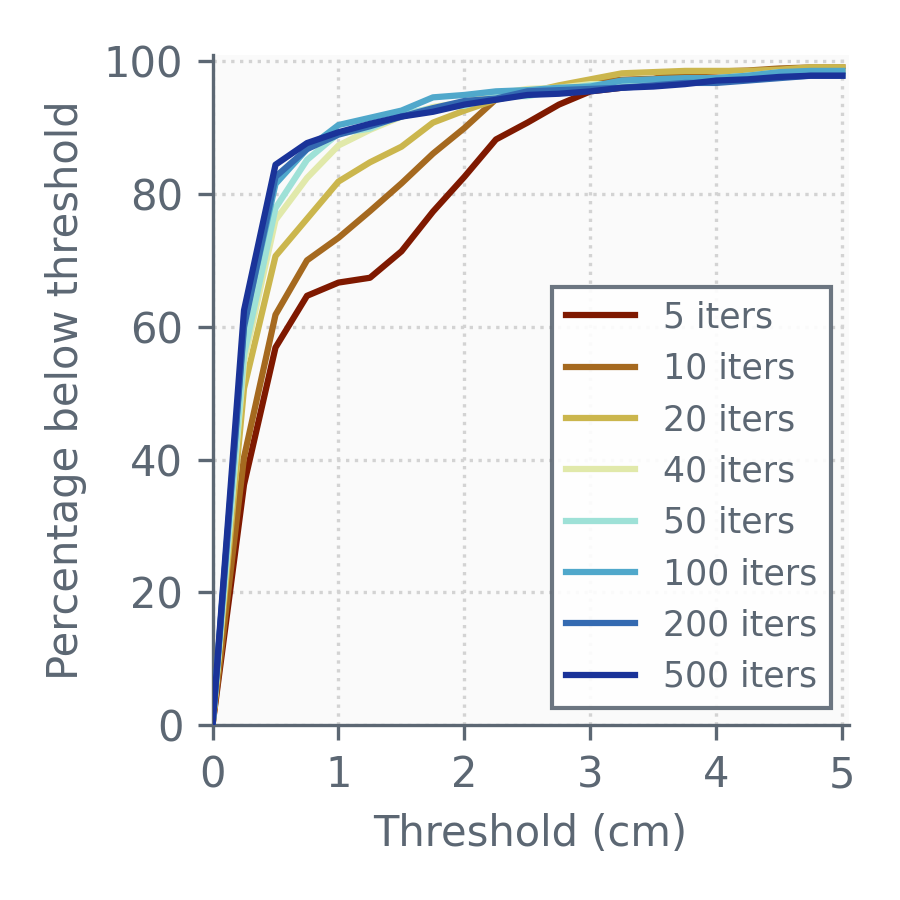} &
        \includegraphics[width=0.24\textwidth]{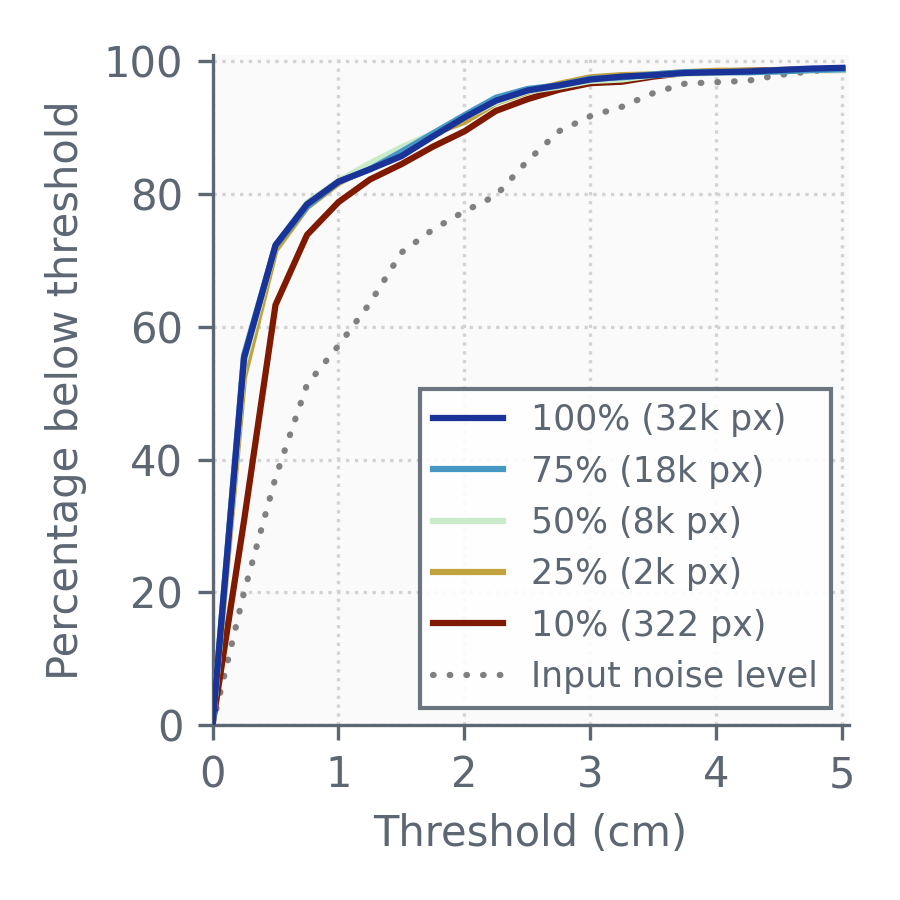}
        
    \end{tabular}
    \caption{
        Left: impact of the number of iterations on Diff-DOPE performance on HOPE~\cite{tyree2022iros:hope}.
        Right: impact of image size on Diff-DOPE performance on HOPE (without the edge matching loss); we also show the mean number of pixels (px) occupied by the object.
    }
    \label{fig:image_size}
\end{figure}

Two additional experiments are shown in Figure~\ref{fig:image_size}.
We varied the number of iterations from 5 to 500 (shown on the left side of Figure~\ref{fig:image_size}), using a reduced HOPE image size (25\% of the original).
Performance is good after a minimum of 50 iterations, and it saturates around 100 iterations. 
With just 5 iterations the AUC is 81.0 compared with 90.5 for 100 iterations.

We also evaluated the impact of image size on performance. 
In our default setting, we use 50\% of the input image size; for HOPE this means reducing 
the image resolution from 1920$\times$1080 to 960$\times$540.
As can be seen from the right side of Figure~\ref{fig:image_size}, the performance does not significantly degrade, even when reduced to 25\% of the original size.
(Note that we omit edge detection in these experiments.)
A negligible decrease in performance is observed between 100\% resolution (87.9 AUC) and 25\% (87.7 AUC);
when the image is decreased to 10\% the AUC falls to 85.3.
At the higher resolutions, the objects occupy an average of 32k pixels down to 2k pixels, which does not cause any appreciable difference in the results.
When reduced to 10\% of the original size, the average object size is just 322 pixels, producing a more noticeable effect.

\subsection{Robot-camera calibration}

\begin{figure}[!t]
    \centering
    \includegraphics[width=0.48\textwidth]{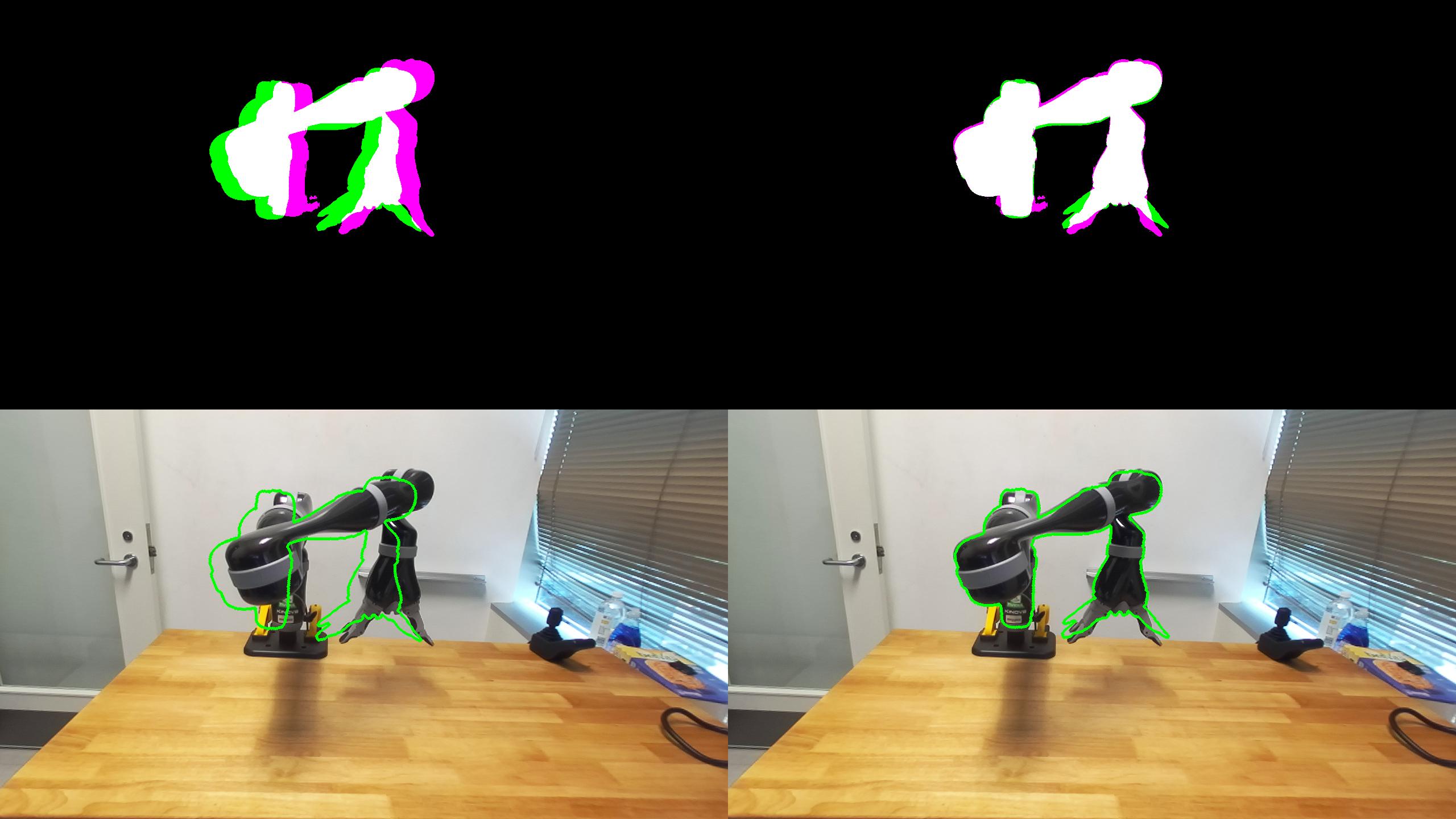} 
    \caption{Diff-DOPE applied to camera-robot calibration.
    Shown are the input pose (left) and output of Diff-DOPE (right).
    Note that the links of the robot match closely, while the errors around the gripper are due to discrepancies in the provided URDF.
    Top: segmentation with green indicating observed, magenta indicating the reprojected 3D model, and white the overlap between the two. 
    Bottom:  The 3D model outline overlaid in green on the observed images. 
    This optimization used two robot configurations captured by a first generation ZED camera. 
    }
    \label{fig:robot}
\end{figure}

While preparing this manuscript we became aware of the recent work of EasyHeC~\cite{chen2023easyhec}, 
which also leverages differentiable rendering for pose estimation, albeit for robot-camera calibration rather than object pose estimation. 
In this section we describe a preliminary experiment showing the potential of using Diff-DOPE to 
calibrate an external camera to a robot in a similar manner. 
While we could have used an existing method, such as DREAM~\cite{lee2020icra:dream}, to get an initial pose for the robot, 
that would require a trained neural network.
Instead we manually estimated a rough initial pose:
Using an RGBD camera (such as ZED), we manually cropped the point cloud to a volume that contained only the robot (assuming no adjacent clutter). 
From the cropped point cloud, we derived a segmentation mask for the original depth image.
Finally, we generated a mesh using the robot's known joint configuration, and manually dragged the mesh into rough alignment with the depth point cloud. %
We repeated this process using a few joint configurations %
to add stability to the optimization. 
Because most URDF robot models have limited texturing, we modified Diff-DOPE to optimize for only segmentation and depth. %
Figure~\ref{fig:robot} shows a qualitative result from running this procedure.
We have successfully used this method in our lab to generate robot-camera calibrations %
that---when combined with Diff-DOPE object poses---allowed for reasonably precise manipulation.\footnote{Initial object poses were predicted by DOPE~\cite{tremblay2018corl:dope}, and segmentation was omitted from Diff-DOPE because DOPE does not predict object masks.}

\section{CONCLUSION}

We have presented a method called Diff-DOPE that leverages a textured 3D model to refine 6-DoF object pose  without any network training. 
The method is based on the idea of differentiable rendering, in which the acquired image is iteratively compared to the rendering of the model until convergence.
Our method leverages RGB, depth, edge detection, and segmentation when available, but it also has flexibility to operate without some of these 
modalities if desired. 
We have examined various parts of the algorithm and shown the importance of \emph{learning rate randomization}, \ie, sampling different learning rates to efficiently optimize the object pose in parallel.

We have shown experiments comparing Diff-DOPE with MegaPose, 
a recent state-of-the-art pose optimizer.
Overall, our method produces more accurate poses without any training, while MegaPose tends to be less disturbed by the initial pose distribution, making it more applicable when the initial pose has more error.
These results support the hypothesis that differentiable rendering has potential to achieve sub-1~cm pose estimation, which is important for robotic grasping and manipulation. 

Future work should include exploring non-Lambertian optimization as part of the pose estimation process. 
This situation is challenging as the textured surface is not well defined, but it will enable pose estimation for objects with more complex surface materials exhibiting reflection and specular highlights.
Even in the mostly matte textured objects used in our experiments, 
 light interacts with the plastic surface of the model during 3D scanning, leading to artifacts that sometimes cause mismatch in the pose estimation process. 
Another promising avenue for research is mixing both classic model-based optimizations and learning-based methods, for which we believe our approach will provide a starting point. 

\section{Acknowledgement}
The authors would like to highlight the indispensable contributions of Jacob Munkberg for code optimization and providing us with CUDA/PyTorch optimized matrix transformations. 
The authors would also like to thank Ankur Handa, Yann Labb\'e, Vladimir Reinharz, and Yen Chen Lin for their feedback during the process of writing this paper.

\bibliographystyle{IEEEtran}
\bibliography{main}

\end{document}